\documentclass[12pt]{l4dc2022}

\title[Control-Tutored Reinforcement Learning]{Control-Tutored Reinforcement Learning: Towards the Integration of Data-Driven and Model-Based Control}
\usepackage{wrapfig}
\usepackage{times}
\usepackage{lipsum}
\usepackage{amsmath}    	
\usepackage{amssymb}	    
\usepackage{amsfonts}		  
\usepackage{mathtools}		
\usepackage{algorithm}		
\usepackage{enumitem}
\usepackage[noend]{algcompatible}

\usepackage{tikz}
\usepackage{pgfplots}
\usepackage{stackengine}
\tikzstyle{block} = [draw, rectangle, 
    minimum height=3em, minimum width=5em]
\tikzstyle{input} = [coordinate]
\tikzstyle{output} = [coordinate]
\tikzstyle{pinstyle} = [pin edge={to-,thin,black}]
\usetikzlibrary{shapes.geometric,shapes.arrows,decorations.pathmorphing,calc}
\usetikzlibrary{matrix,chains,scopes,positioning,arrows,fit}

\usepackage{bm}           

\setlist{noitemsep} 		          

\newcommand{\T}{^{\mathsf{T}}}

\newcommand{\B}[1]{\if#1\relax\bm{#1}\else\mathbf{#1}\fi} 
\newcommand{\R}[1]{\mathrm{#1}}						      
\newcommand{\C}[1]{\mathcal{#1}}
\newcommand{\BB}[1]{\mathbb{#1}}
\newcommand{\norm}[1]{\left\lVert #1 \right\rVert}

\usepackage[prependcaption,colorinlistoftodos]{todonotes}

\author{%
 \Name{Francesco De Lellis} \Email{francesco.delellis@unina.it}\\
 \addr University of Naples Federico II, Italy%
 \AND
 \Name{Marco Coraggio} \Email{marco.coraggio@unina.it}\\
 \addr Scuola Superiore Meridionale, Italy%
 \AND
 \Name{Giovanni Russo$^*$} \Email{giovarusso@unisa.it}\\
 \addr University of Salerno, Italy%
 \AND
 \Name{Mirco Musolesi$^*$} \Email{m.musolesi@ucl.ac.uk}\\
 \addr University College London, UK, and University of Bologna, Italy%
 \AND
 \Name{Mario di Bernardo$^*$} \Email{mario.dibernardo@unina.it}\\
 \addr University of Naples Federico II, Italy, and Scuola Superiore Meridionale, Italy%
}

\begin{document}

\maketitle

\begin{abstract}%
We present an architecture where a feedback controller derived on an approximate model of the environment assists the learning process to enhance its data efficiency.
This architecture, which we term as  Control-Tutored Q-learning (CTQL), is presented in two alternative flavours. 
The former is based on defining the reward function so that a Boolean condition can be used to determine when the control tutor policy is adopted, while the latter, termed as probabilistic CTQL (pCTQL), is instead based on executing calls to the tutor with a certain probability during learning.
Both approaches are validated, and thoroughly benchmarked against Q-Learning, by considering the stabilization of an inverted pendulum as defined in OpenAI Gym as a representative problem.
\end{abstract}



\vspace{0.5cm}
\section{Introduction}
Reinforcement learning (RL) is a popular framework for the control of autonomous agents, due to its ability to autonomously derive control policies without assuming that the dynamics of the environment are known \citep{Sutton1998,bertsekas1996neuro}.
Despite the many remarkable successes in different applications of this type of approaches \citep{nian2020review}, two key problems for RL algorithms remain to be solved: (i) potentially {\em long} learning times and (ii) the lack of convergence or performance guarantees (important for example in safety-critical applications) during learning \citep{berkenkamp2017safe,pfeiffer2018reinforced}. 

To overcome these problems, a possible solution, particularly for control applications of RL, is to adopt model-based solutions where the learning agent derives and refines a data-driven model of the environment during the learning process. Examples in the literature include \cite{deisenroth2011pilco,kurutach2018model} among many others. 
However, in many control applications, some equation-based models of the environment are often available, even though they might not be accurate enough to allow for an entirely model-based solution of the control problem. When classical RL is used, such approximate or partial models of the environment are often discarded in favour of a completely model-free approach. 

In this paper, we investigate the possibility of embedding a feedback control law synthesized by using a partial model of the environment to assist the learning process.
In the same spirit as human-assisted learning strategies where data collected from {\em humans} in the loop are exploited to enhance the learning process \citep{lien2009interactive,suppakun2020coaching,zhan2021human,nguyen2019apprenticeship}, here we propose the use of a {\em feedback controller} in the loop with the aim of steering the learning phase, reducing the amount of data samples required and improving the ability of learned policies to achieve a given control goal.
The contributions of this paper can be summarized as follows: (1) we propose a novel algorithm that leverages the use of a feedback controller in the loop to make the RL process more data efficient; (2) we present both a deterministic and a probabilistic approach to implement the strategy above and decide when assistance from the control tutor (the feedback controller in the loop) is invoked during the learning; (3) by using a set of aptly defined metrics, we compare the performance of the novel approaches with those of a classical RL algorithm both from a learning and a control perspective by using the inverted pendulum benchmark implementation from OpenAI Gym~\citep{brockman2016openai}.


We wish to emphasize that, to our knowledge, this is the first analysis of this type for algorithms where learning is assisted via a feedback controller. Our results convincingly show that such ``control-tutored'' learning approaches require fewer data samples and/or obtain  higher rewards, while achieving smaller errors in control regulation tasks.
    

\section{Related Work}
Several solutions in the existing literature aim at combining control theoretic strategies with reinforcement learning to solve control problems. 
In particular, various approaches combine RL with model predictive control (MPC).
For instance, in \cite{rathi2020driving}, a MPC is used to decide the action when the state of the system to control is in a certain region, while the action taken from a $Q$-table is used otherwise; the table being updated after every action.
The use of a (linear) MPC strategy is again suggested in \cite{zanon2020safe}, where a reinforcement learning module can vary the parameters of the cost function and refine the available model of the system to control.

Other solutions combining control strategies with RL include those in \cite{abbeel2006using}, where a policy gradient algorithm is adopted which uses preexisting knowledge of the system dynamics in the form of an approximate Markov decision process; or that presented in \cite{li2021safe}, where a \emph{reference action governor} is used to enforce safety constraints (in the sense of restricting the state space to admissible regions). In so doing, the action is decided via an optimization problem that penalizes deviations from the action suggested by a RL strategy, making these approaches a valuable solution to achieve safe RL.

A strategy similar in spirit to the CTQL we propose here is contained in \cite{argerich2020tutor4rl}, where a Deep $Q$-Network is extended with the policy having a probability to take an action dictated by an ``expert'', which can solve the control problem, to improve data efficiency. 
However, differently from \cite{argerich2020tutor4rl}, in our CTQL algorithm, we consider the ``expert'' to be a feedback control law, that if deployed on its own would be unable to achieve the control goal.
Also, note that contrary to previous approaches, e.g. \cite{deisenroth2011pilco}, where an approximation for the system dynamics is learned during the control steps, here we assume to possess and exploit some information on the environment model  before simulations so as to be able to derive some feedback control law to be used to assist the learning phase.
An earlier preliminary version of the CTQL was recently presented in \cite{de2020tutoring}.
\section{Mathematical Preliminaries}
%
\paragraph{Notation.}
Sets are denoted by calligraphic capital characters and random variables are denoted via capital letters. For example, $X$ is a random variable and we denote its realization by $x$.
The probability density (mass) function of the continuous (discrete) random variable $X$ is denoted by $p(x)$ and we use the notation $x\sim p(x)$ to denote the sampling of a random variable from its probability function.
For both continuous and discrete random variables, we always consider the situation where the support of $p(x)$ is compact;
$\R{rand}(\C{A})$ denotes the uniform distribution over the set $\C{A}$.
The expectation of a function, say $h(\cdot)$, of $X$ is defined as
$\BB{E}_p[h(X)] \coloneqq \int h(X)p(x) \R{d}x$, when this is continuous; if $X$ is discrete, we have $\BB{E}_p[h(X)] \coloneqq \sum h(x)p(x)$.
In both cases, the integral/sum is taken on the support of $p(x)$, and we might omit $p$ in $\BB{E}_p$ when there is no ambiguity. 
We denote by $\norm{\cdot}$ the Euclidean norm.

\paragraph{Problem set-up.}
We consider a discrete time dynamical system affected by noise, of the form
\begin{equation}\label{eq:dynamical_system}
    X_{k+1} = f_k(X_k, U_k, W_k),
    \quad x_0 = \tilde{x}_0,
\end{equation}
where $k \in \BB{N}_{\ge 0}$ is discrete time,
$X_k \in \C{X}$ is the {state} of the system at time $k$, with
$\C{X}$ being the state space, 
$\tilde{x}_0 \in \C{X}$ is the initial condition, 
$U_k \in \C{U}$ is the {control input} (or {action}) and $\C{U}$ is the set of feasible inputs. Also,  $W_k$ is a random variable representing {noise} and $f_k : \C{X} \times \C{U} \times \C{W} \rightarrow \C{X}$ is the system's {dynamics}.


Following e.g. \cite{matni2019self,recht2019tour}, given this set-up, we consider the problem of learning a plan of actions {$\pi_1, \dots, \pi_{N-1}$} to solve the following finite-horizon optimization problem: 
\begin{subequations}\label{eq:rl_problem_objective}
\begin{align}
    \max_{\pi_1,\ldots,\pi_{N-1}}& \ \ \BB{E}[J^{\bar{\pi}}],\\ 
    \text{s.t.}
    & \ \ X_{k+1} = f_k(X_k,U_k,W_k),
        \quad k \in \{ 1, \dots, N-1 \},\\
    &\ \ U_k = \pi_k(X_{k}),  
        \quad k \in \{ 0, \dots, N-1 \},\\
    &\ \ x_0 \ \text{given},
\end{align}
\end{subequations}
where the time horizon is between $0$ and $N$. In (\ref{eq:rl_problem_objective}) the cost is set as the expectation of the \emph{objective function}
\begin{equation} \label{eq:objective}
    J^{\bar{\pi}} = r_N(X_{N}) + \sum_{k=1}^{N-1} r_k(X_{k}, X_{k-1}, U_{k-1}),
\end{equation}
with $r_k : \C{X} \times \C{X} \times \C{U} \rightarrow \BB{R}$ and $r_N:\C{X} \rightarrow \BB{R}$ being the \emph{rewards} received, at each $k$, by the agent. In what follows,
whenever we assume a function or quantity is stationary, we drop the subscript $k$ in the notation.


We observe that in many RL scenarios, even if the system dynamics $f_1, \dots, f_{N-1}$ are not perfectly known, some partial knowledge about the plant (from e.g. first-principles) might be available and encoded in some mathematical model of the plant. We propose that this limited information can be exploited to design a feedback control law (or control tutor) that can be used to assist and drive the learning process towards the solution of a control problem of interest, reducing the learning times and improving the control performance. In particular, the control tutor can be invoked under certain circumstances during the learning stage to suggest actions that the agent can take as an alternative to those computed using a more traditional approach, e.g. obtained by reading the $Q$-table. 
%


\section{Control Tutored Reinforcement Learning}


%
\begin{figure}
\centering
\includegraphics[width=0.5\textwidth]{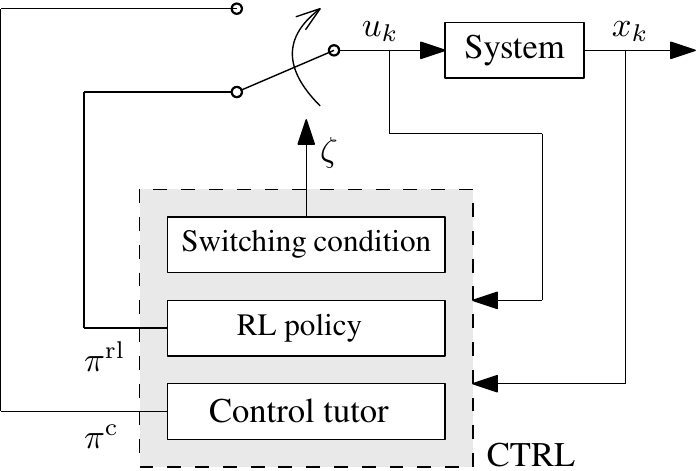}
\caption{Schematic of the Control-Tutored Reinforcement Learning (CTRL) framework.}
\label{fig:ctrl_scheme}
\end{figure}

We start by assuming that we have an estimate of $f$, say $\hat{f} : \C{X} \times \C{U} \rightarrow \C{X}$, so that the dynamics of system \eqref{eq:dynamical_system} is rewritten as
$f(x,u,w) = \hat{f}(x, u) + \delta (x, u, w)$,  $\forall x \in \C{X}, \forall u \in \C{U}, \forall w \in \C{W}$,
where $\delta : \C{X} \times \C{U} \times \C{W} \rightarrow \C{X}$ describes the effect of unknown terms in the dynamics and/or of noise on the system's dynamics. We term  the (possibly, model-based) policy obtained by considering only $\hat{f}$ as the \emph{control tutor policy}, and denote it by $\pi^{\R{c}} : \C{X} \rightarrow \C{U}$.

The architecture of the \emph{Control Tutored Reinforcement Learning} (CTRL) \citep{de2020tutoring} is schematically shown in Figure \ref{fig:ctrl_scheme}. 
The figure highlights the presence of a switching condition $\zeta$ that orchestrates, at each $k$, the use of either a policy coming from a RL algorithm or the tutor policy. The result is the following switching policy used for learning: 
\begin{equation}\label{eq:ctrl}
    \pi(x) =
    \begin{dcases}
        \pi^{\R{rl}}(x), & \text{if } \zeta \ \text{is true,}\\
        \pi^{\R{c}}(x), & \text{otherwise},
    \end{dcases}
\end{equation}
where $\zeta$ is a Boolean function (that might depend on time, previous states, etc.) and $\pi^{\R{rl}}$ is the policy of a RL algorithm.

For concreteness, we now provide a simple expression for the control tutor policy $\pi^{\R{c}}$.
First, let $\bar{\C{U}} \supseteq \C{U}$ ($\bar{\C{U}}$ might be a continuous set whose discretization yields $\C{U}$); then, from $\hat{f}$ we can design a feedback control strategy $v : \C{X} \rightarrow \bar{\C{U}}$.
At this point, from $v$, letting $\epsilon^{\R{c}} \in (0, 1)$, and $\forall x \in \C{X}$, we take the \emph{control tutor policy} in \eqref{eq:ctrl} as
\begin{equation}\label{eq:tutor_policy}
    \pi^\R{c}(x) = \begin{dcases}
        \arg \min\limits_{ u \in \C{U}} \norm{v(x) - u},
            & \text{with probability } 1 - \epsilon^{\R{c}},\\
        u \sim \R{rand}(\C{U}),
            & \text{with probability } \epsilon^{\R{c}},\\
    \end{dcases}
\end{equation}
On the other hand, for the reinforcement learning policy $\pi^{\R{rl}}$ in \eqref{eq:ctrl}, we adopt an $\epsilon$-greedy Q-learning solution, i.e.,
\begin{equation}\label{eq:rl_policy}
    \pi^{\R{rl}}(x_k) = \begin{dcases}
        \arg \max_{u\in \C{U}}Q_k(x_k,u),
            & \text{with probability } 1 - \epsilon^{\R{rl}},\\
        u \sim \R{rand}(\C{U}),
            & \text{with probability } \epsilon^{\R{rl}},\\
    \end{dcases}
\end{equation}
where $\epsilon^{\R{rl}} \in (0, 1)$ and $Q_k : \C{X} \times \C{U} \rightarrow \BB{R}$ is the well-known \emph{state-action value function} \citep{Sutton1998, bertsekas1996neuro} at time $k$.

At time $k$, once an action is selected from either $\pi^{\R{rl}}$ or $\pi^{\R{c}}$, the corresponding reward is obtained and used to update the $Q$-table according to the law
\begin{equation}\label{eq:Q_update}
    Q_{k+1}(x_{k},u_{k}) = (1-\alpha)Q_k(x_{k},u_{k}) + \alpha[r(x_{k+1}, x_{k}, u_{k}) + \gamma \max\limits_{u \in \C{U}}Q_k(x_{k+1},u)],
\end{equation} 
where $\alpha \in (0, 1]$ is the \emph{learning rate} and $\gamma \in (0, 1]$ is the \emph{discount factor}.

The remaining term to be defined in \eqref{eq:ctrl} is $\zeta$.
In the following, we present two alternative choices for $\zeta$ that result into two different algorithms.

\subsection{Control-Tutored Q-Learning}\label{sec:ctql}

This first algorithm based on the CTRL framework is the \emph{Control-Tutored Q-learning} (CTQL), which was first presented in \cite{de2020tutoring}.
This algorithm uses a reward with a specific structure.
In particular, let $x^* \in \C{X}$ be a \emph{goal state}, 
and $d(x) \coloneqq \norm{x - x^*}^2$.
Moreover, letting $\theta \in \BB{R}_{>0}$ and $\bar{\rho} \in \BB{R}_{>0}$, we let the prize function
\begin{equation}\label{eq:prize_reward}
    \rho(x) = \begin{dcases}
    \bar \rho, & \text{if } \norm{x - x^*} < \theta, \\
    0, & \text{otherwise},
    \end{dcases}
\end{equation}
and the reward $r_k$ in \eqref{eq:objective} is given as
\begin{equation}\label{eq:reward_ctql}
    r(X_{k}, X_{k-1}, U_{k-1}) = d(X_{k-1}) - d(X_{k}) + \rho(X_{k}), 
    \quad k=1,\ldots,N-1,
\end{equation}
with $r_N(X_N) = 0$.
The switching criterion $\zeta$ in \eqref{eq:ctrl} depends on the current state $x_k$, where
\begin{equation}\label{eq:zeta_ctql}
    \zeta \ \text{is} \begin{cases} 
    \text{true},  & \text{if}\ \max\limits_{u \in \C{U}} Q_k(x_k,u) > 0,\\
    \text{false}, & \text{otherwise}.
    \end{cases}
\end{equation}
Additionally, $\forall x \in \C{X}, \forall u \in \C{U}$, we initialize $Q_0(x,u) = 0$.
Thus, in the first phase of learning, when limited information about the environment is available, the control tutor policy $\pi^{\R{c}}$ drives the learning process.
Then, gradually, as the values of the $Q$-table are updated using \eqref{eq:Q_update}, the reinforcement learning policy $\pi^{\R{rl}}$ is preferred.
%
%
%
\subsection{Probabilistic Control-Tutored Q-Learning}
Although we found the CTQL to have better performance with respect to the classical Q-learning in certain scenarios (see Section \ref{sec:comparison_learning_performance}), the reward \eqref{eq:reward_ctql} does not satisfy the hypotheses used in the classical proof of convergence used for the Q-learning (see, e.g., \citep{bertsekas1996neuro}), as it is not either non-negative or non-positive.

Moreover, we verified that
the CTQL fails when the reward function is changed or not chosen appropriately.
Therefore, we propose next a simpler probabilistic-based choice for the Boolean condition
$\zeta$ in \eqref{eq:ctrl}. We name the resulting algorithm as \emph{probabilistic Control Tutored Learning} (pCTQL), where by removing the constraints on the reward function required by the deterministic approach, we can use a more standard reward function (compliant with e.g. the classic QL proofs).

In particular, letting $\beta \in [0, 1]$,
\begin{equation}\label{eq:zeta_pctql}
    \zeta \ \text{is} \begin{cases} 
    \text{true},  & \text{with probability } \beta,\\
    \text{false}, & \text{otherwise},
    \end{cases}
\end{equation}
the pCTQL policy is defined as
\begin{equation}\label{eq:policy_pctql}
    \pi(x) = \begin{dcases}
        \arg \max_{u\in \C{U}}Q_k(x,u),
            & \text{with probability } \beta (1 - \epsilon^{\R{rl}}),\\
        \arg \min\limits_{ u \in \C{U}} \norm{v(x) - u},
            & \text{with probability } \omega \coloneqq (1 - \beta) (1 - \epsilon^{\R{c}}),\\
        u \sim \R{rand}(\C{U}),
            & \text{otherwise},\\
    \end{dcases}
\end{equation}
Note that it is also possible to introduce a dependency of the probability $\beta$ on the current state, time, or other quantities.


\section{Metrics}
Here we define several metrics to characterize and compare quantitatively the performance of different control algorithms. Each numerical simulation is run in $S \in \BB{N}_{> 0}$ independent sessions.
Each session is composed of $E \in \BB{N}_{> 0}$ episodes: the learned quantities (e.g., $Q$-table) are carried over from one episode to the next, and re-initialized at each session.
Each episode consists of a simulation of $N \in \BB{N}_{> 0}$ time steps.
We let $J_e^{\pi}$ be the cumulative reward (as given in \eqref{eq:objective}) obtained in episode $e$.
Moreover, we let the \emph{goal condition} be a Boolean proposition that assesses whether the control goal can be considered as having been achieved in an episode (the specific form of the goal condition depends on the task at hand). 
We define the following three metrics to assess the learning performance.
\begin{definition}[Learning metrics]\label{def:learning_metrics}
    (i) The \emph{average cumulative reward} is
    $J_{\R{avg}}^{\pi} \coloneqq \frac{1}{E} \sum_{e = 1}^E J_e^{\pi}$.
    (ii) The \emph{terminal episode} $E_{\R{t}}$ is the smallest episode such that the goal condition is satisfied for all $e \in \{E_{\R{t}}- 30, \dots, E_{\R{t}}\}$.
    (iii) The \emph{average cumulative reward after terminal episode} is
    $J_{\R{avg}, \R{t}}^{\pi} \coloneqq \frac{1}{E_{\R{t}}} \sum_{e = E_{\R{t}}}^E J_e^{\pi}$.
\end{definition}
$J_{\R{avg}}^{\pi}$ is a common metric typically used in RL \citep{duan2016benchmarking, wang2019benchmarking}; $E_{\R{t}}$ is used to assess when the learning phase might be considered concluded, and thus to evaluate data efficiency; $J_{\R{avg}, \R{t}}^{\pi}$ describes how performing the controller is, in terms of rewards, once training is completed.



Next, we define two metrics inspired by those commonly used in control theory to assess the transient and steady-state performance of an algorithm.
Let again $x^*$ be a goal state, let $\eta \in \BB{R}_{\ge 0}$, $N^- \in \BB{N}_{>0}$ with $N^- < N$,  and let the goal condition be true if
\begin{equation}\label{eq:goal_condition}
    \exists \bar{k} \in [0, N^-] : 
    \norm{x_k - x^*} \le \eta, 
    \quad \forall k \in [ \bar{k}, N].
\end{equation}
\begin{definition}[Control metrics]\label{def:control_metrics}
        (i) In an episode, the \emph{settling time} $k_{\R{g}}$ is the smallest value of $\bar{k}$ that fulfills \eqref{eq:goal_condition}.
        (ii) The \emph{steady state error} is $e_{\R{g}} \coloneqq \frac{1}{N - k_{\R{g}} + 1} \sum_{k = k_{\R{g}}}^{N}
        \norm{x - x_{\R{g}}}$.
\end{definition}
%
%
\section{Benchmark Description}

\subsection{Control Problem}
As a benchmark problem to compare the performance of the proposed algorithms, we consider the problem of stabilizing a pendulum in its inverted position, provided by the OpenAI Gym framework  \citep{brockman2016openai, GYM}.
This problem is particularly representative for two reasons. 
(i) As the upward position is unstable and the the system dynamics is nonlinear, this problem is typically used in control theory as a test for new control strategies \citep{Khalil:1173048}.
(ii) We will select a linear feedback controller ($v$ in \eqref{eq:tutor_policy}), which by itself cannot stabilize the pendulum.
This means that any benefit observed when using CTQL and pCTQL will be due to the combination of the reinforcement learning policy and the model-based one, and not just the latter.

\paragraph{Environment.}
The pendulum is a rigid rod of length $l = 1 \ \R{m}$, with a homogeneous distribution of mass $m = 1 \ \R{kg}$;
its moment of inertia is $I = ml^2/3$ and it is affected by gravity, with acceleration $g$.
We let $x_k = [x_{1,k} \ \ x_{2,k}]\T$, where $x_{1,k}$ and $x_{2,k}$ are the angular position and angular velocity of the pendulum, respectively; $x_{k,1} = 0$ corresponds to the unstable vertical position.
The control input $u_k$ is a torque applied to the pendulum.
The discrete-time dynamics is obtained by discretizing the continuous-time dynamics with a sampling time $T = 0.05 \ \R{s}$ using the forward Euler method.
Unless noted otherwise, the initial condition is the downward stable position $\tilde{x}_0 = [\pi \ \ 0]\T$.
\paragraph{State and control spaces.}
The spaces for states and control variable are bounded, so that
$x_{k} \in \left[ -{\pi}, {\pi} \right] \times \left[-8, 8 \right]$, and
$u_k \in [-2, 2]$.
Additionally, both spaces are discretized as follows.
Concerning $x_{1, k}$, 
the interval $\left[ -\pi, -\frac{\pi}{9} \right]$ is discretized into 8 equally spaced values,
$\left( -\frac{\pi}{9}, -\frac{\pi}{36} \right]$ into 7 values, and
$\left( -\frac{\pi}{36}, 0 \right]$ into 5 values;
$\left[ 0, \pi \right]$ is discretized in an analogous fashion.
Concerning $x_{2, k}$, 
$\left[ - 8, -1 \right]$ is discretized into 10 values, and
$\left( - 1, 0 \right]$ into 9 values
(analogously for $\left[ 0, 8 \right]$).
Concerning $u_k$, 
$\left[ - 2, -0.2 \right]$ is discretized into 9 values, and
$\left( -0.2, 0 \right]$ into 4 values
(analogously for $\left[ 0, 2 \right]$.
%
\paragraph{Iterations.}
For each set-up, we run $S = 10$ sessions and average the results.
For each session, we run $E = 10000$ episodes, composed of $N = 400$ time steps.
%
\paragraph{Goal and rewards.}
The objective is to stabilize the pendulum in its upward position, $x^* = [0 \ \ 0]\T$.
Concerning the goal condition in \eqref{eq:goal_condition}, we take $N^- = 300$ and $\eta = 0.05 x_{\R{max}}$, where $x_{\R{max}} \coloneqq \norm{[\pi \ \ 8]}$.
This goal is encoded in two reward functions.
The first one is 
\begin{equation}\label{eq:reward_distance}
    r^{\R{a}}(X_{k}, X_{k-1}, U_{k-1}) = d(X_{k-1}) - d(X_{k}) + \rho(X_{k}),
\end{equation}
where $d(x) \coloneqq x_{1}^2 + 0.1 x_{2}^2$, and 
$\rho$ was given in \eqref{eq:prize_reward}, with $\bar{\rho} = 5$ and $\theta = 0.05$.

The second reward function we will consider is the standard Gym reward, i.e.,
\begin{equation}\label{eq:reward_gym}
    r^{\R{g}}(X_{k}, X_{k-1}, U_{k-1}) = X_{1,k}^2 + 0.1 X_{2,k}^2 + 0.001 U_{k-1}^2.
\end{equation}
%
\paragraph{Hyperparameters.}
In \eqref{eq:Q_update}, we take $\gamma = 0.97$ and $\alpha = \left( 1+\frac{e}{1000} \right)^{-\frac{1}{2}}$, where $e$ is the current episode \citep{even2003learning}, so that the learning rate decays approximately from $0.7$ to 
$0.3$, over $10000$ episodes.
In \eqref{eq:tutor_policy} and \eqref{eq:rl_policy}, we take $\epsilon^{\R{c}} = \epsilon^{\R{rl}} = 0.03$.
Concerning $\beta$ in \eqref{eq:zeta_pctql}, we tested $ \beta \in \{ 0.9990 , 0.9948, 0.9897, 0.9485, 0.8969 \}$, which approximately corresponds to $\omega \in \{ 0.001, 0.005, 0.010, 0.050, 0.100 \}$ in \eqref{eq:policy_pctql}. 
%
\paragraph{Feedback control law.}
We assume we have partial information on the pendulum dynamics, in the form of an approximate dynamics $\hat{f}$.
In particular, $\hat{f}$ is the linear dynamics that is topologically equivalent to the nonlinear dynamics of the pendulum, close to the origin $[0 \ \ 0]\T$ (also the goal state).
Namely, 
$\hat{f}\left(x_{k}, v_{k}\right) = A x_{k} + B v_{k}$,
where 
$A = \left[ \begin{smallmatrix} 
0  &  1 + T  \\
3 T g /2 l  &  1
\end{smallmatrix} \right]$
and 
$B = \left[ \begin{smallmatrix} 
0   \\
T/I
\end{smallmatrix} \right]$
.
From $\hat{f}$, we synthesize the linear controller $v_k = -Kx_k$, where $K = [5.83 \ \ 1.83]\T$.
This controller can locally stabilize the pendulum in its inverted position from nearby initial conditions, and is obtained, for the sake of simplicity, via a pole placement technique, assigning poles 
to have an acceptable settling time.
Note that this controller if used on its own is in unable to swing up the pendulum from its downward asymptotically stable position.

\section{Comparison of Learning Performance}
\label{sec:comparison_learning_performance}
\begin{figure}[t]
\centering
\stackunder[1pt]{\includegraphics[width=.48\linewidth]{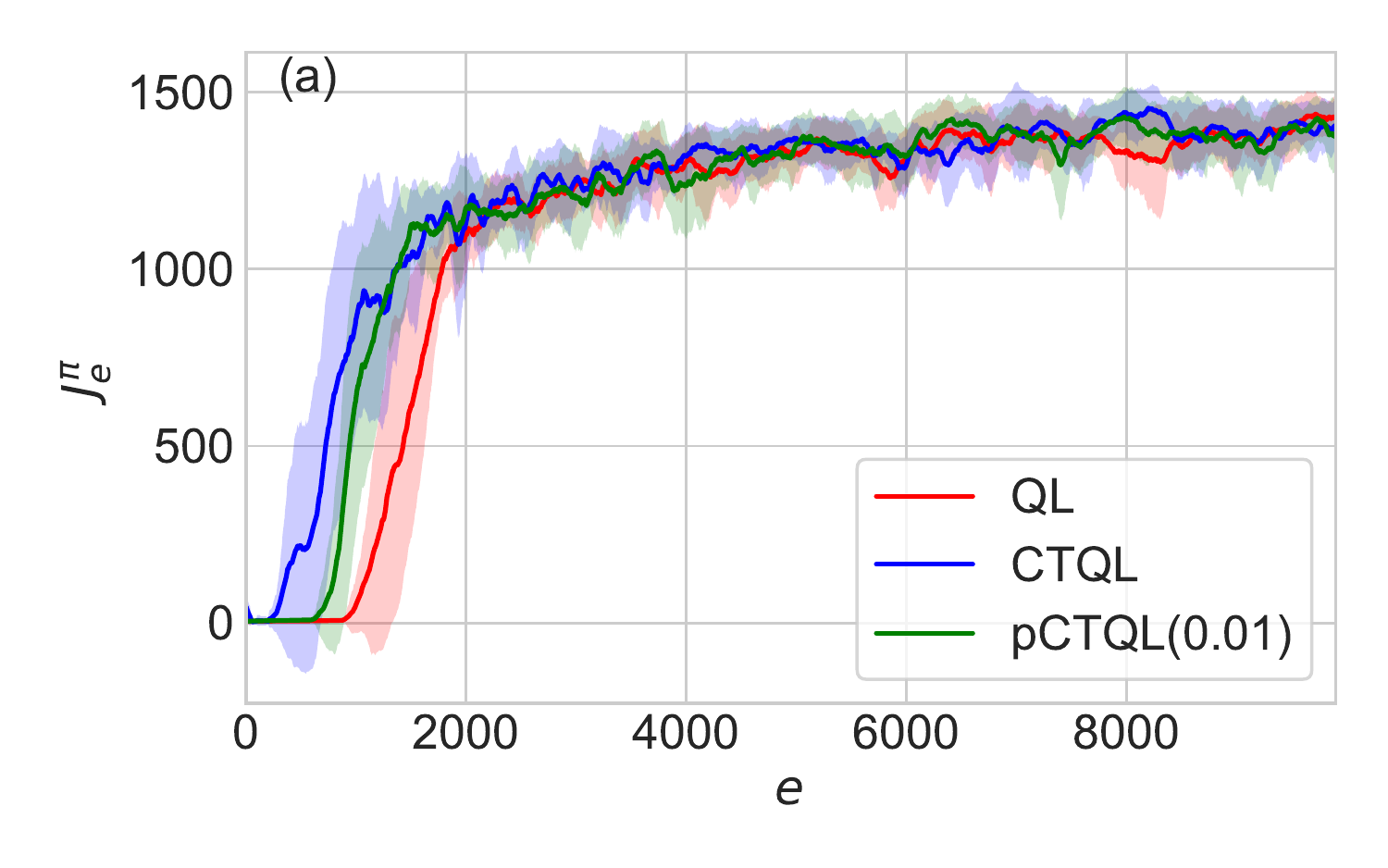}}{}
\label{fig:scenario_1_reward}
\stackunder[1pt]{\includegraphics[width=.48\linewidth]{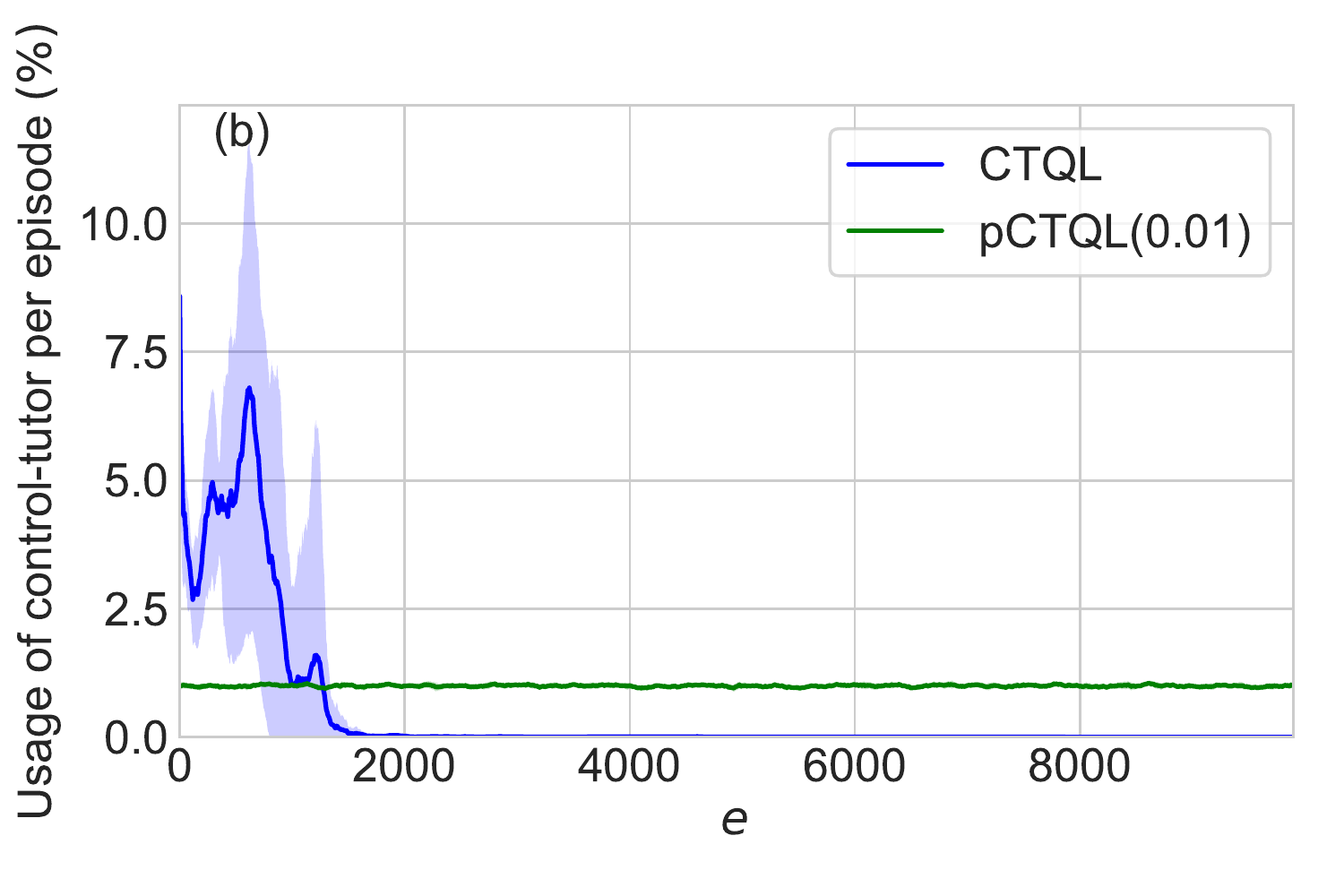}}{}
\label{fig:scenario_1-tutor}
\caption{(a) Cumulative reward per episode $J^\pi_e$, obtained with reward \eqref{eq:reward_distance}. 
(b) Percentage of steps the control-tutor policy $\pi^{\R{c}}$ was used in each episode. 
In both (a) and (b) the solid curves are the mean of the results of $S$ sessions; for readability, the curves are averaged with a moving average of 100 samples (taken on the right); shaded areas corresponds to the means plus or minus the standard deviations.}
\label{fig:scenario_1}
\end{figure}

\begin{figure}[t]
\centering
\includegraphics[width=1\linewidth]{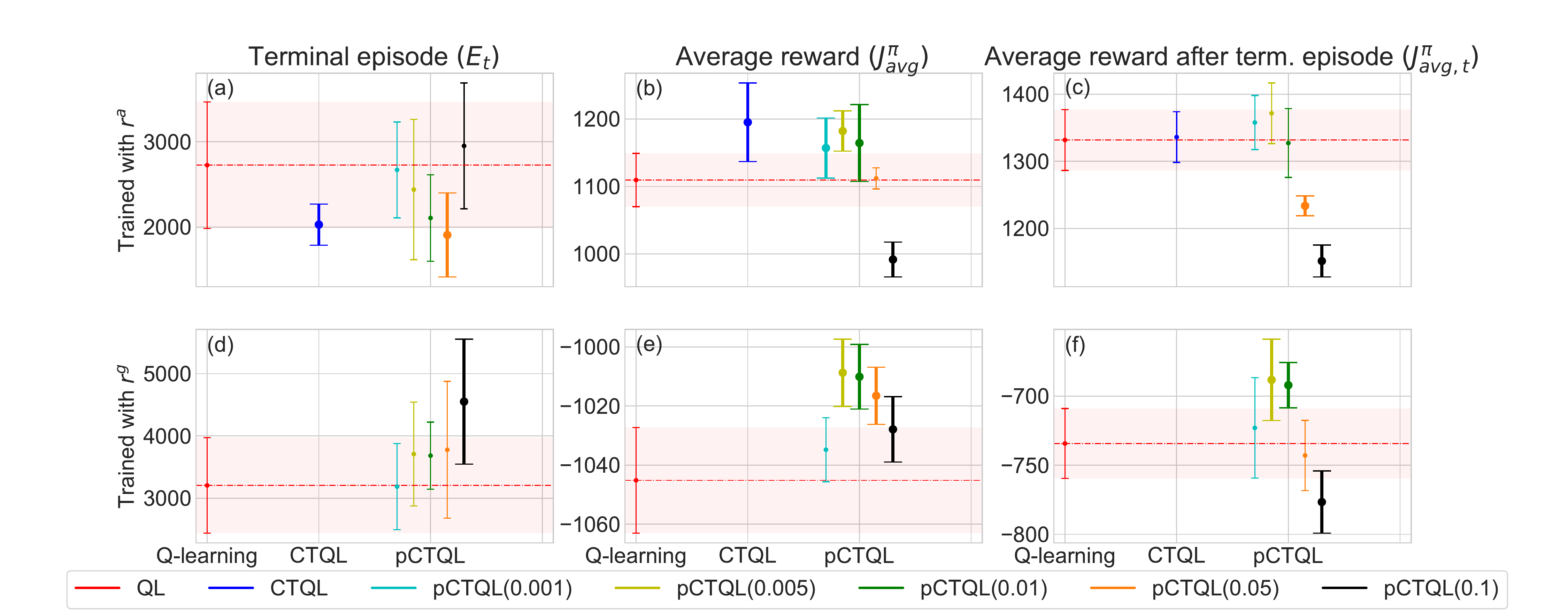}
\caption{(a), (b), (c): $E_{\R{t}}$, $J_{\R{avg}}^{\pi}$, and $J_{\R{avg}, \R{t}}^{\pi}$, respectively (Definition \ref{def:learning_metrics}), with reward \eqref{eq:reward_distance}.
(d), (e), (f): analogously but with reward \eqref{eq:reward_gym}.
pCTQL is reported with different values of $\omega$, expressed in percentage.
The means and standard deviations of $S$ sessions are portrayed. 
Values that are statistically significantly different from those of the Q-learning are in bold (according to a Welch’s t-test with $p$-value less than $0.05$ \cite{welch1947generalization}).}
\label{fig:learning_metrics}
\end{figure}

%
%
\paragraph{Case of reward (\ref{eq:reward_distance}).}
First, we compare Q-Learning, CTQL and pCTQL with different values of $\omega$, when using reward \eqref{eq:reward_distance}. 
The results are reported in Figures
\ref{fig:scenario_1}.(a)--(b), and \ref{fig:learning_metrics}.(a)--(c), containing the cumulative reward per episode $J_e^\pi$, 
the frequency with which the control tutor is used, and the learning metrics (Definition \ref{def:learning_metrics}), respectively.
For the sake of clarity, in Figure \ref{fig:scenario_1} the results of the pCTQL were only plotted for $\omega = 0.01$, as we found that value to give the best performance overall.

From Figure \ref{fig:learning_metrics}.(a), comparing CTQL and pCTQL to Q-learning, we observe that $E_{\R{t}}$---a measure of data efficiency---is smaller (by a statistically significant margin) for the CTQL and for the pCTQL with $\omega = 0.05$; on the other hand, the pCTQL with other values of $\omega$ are on par with the Q-learning.
This fact is also visible in Figure \ref{fig:scenario_1}.(a), as the reward curves of pCTQL and CTQL grow earlier than that of Q-learning, and in Figure \ref{fig:learning_metrics}.(b), showing that the average rewards $J_{\R{avg}}^{\pi}$ are higher for CTQL and pCTQL.
Finally, Figure \ref{fig:scenario_1}.(b)  shows that CTQL uses the control tutor policy more in the beginning, and progressively less as episodes are completed.

\paragraph{Case of reward (\ref{eq:reward_gym})}

We also compared the performance of Q-learning and pCTQL when using reward \eqref{eq:reward_gym}; the results are portrayed in Figures 
\ref{fig:learning_metrics}.(d)--(f).
We see that pCTQL with $\omega = 0.01$ is comparable to Q-learning in terms of learning time ($E_{\R{t}}$), yet obtains a larger average reward ($J_{\R{avg}}^{\pi}$) and average reward after terminal episode ($J_{\R{avg}, \R{t}}^{\pi}$), confirming the effectiveness of a control tutor-based architecture, even when the reward has a structure different from \eqref{eq:reward_ctql}.
\begin{figure}[t]
\centering
\includegraphics[width=1\linewidth]{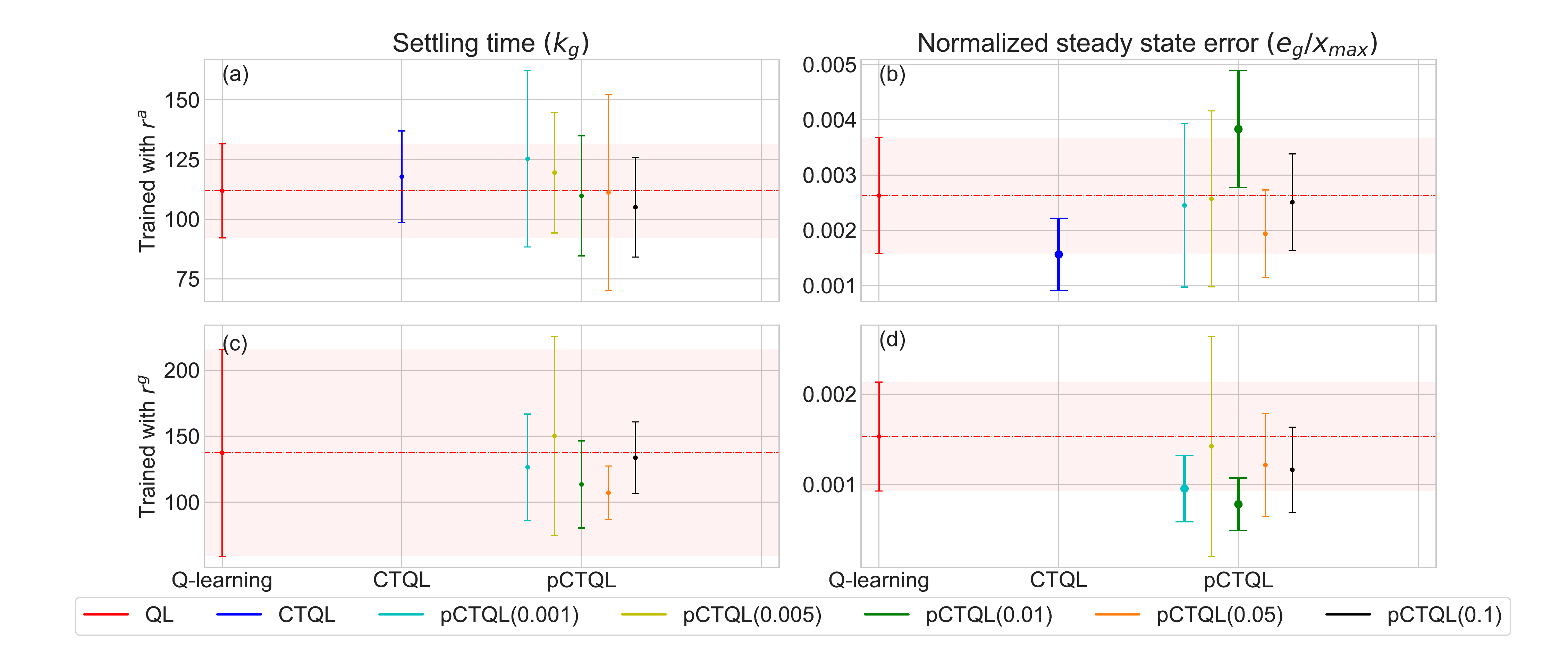}
\caption{(a), (b): $k_{\R{g}}$ and $e_{\R{g}} / x_{\R{max}}$, respectively (Definition \ref{def:control_metrics}), with reward \eqref{eq:reward_distance}, under nominal conditions.
(c), (d): analogous, but with reward \eqref{eq:reward_gym}.}
\label{fig:validation_metrics}
\end{figure}

\section{Comparison of Control Performance}
%
\paragraph{Nominal conditions.}
We also compared the algorithms in terms of their control performance at the end of the learning stage, using the metrics given in Definition \ref{def:control_metrics}.
The results, using both rewards \eqref{eq:reward_distance} and \eqref{eq:reward_gym} are shown in Figure \ref{fig:validation_metrics}.
Firstly, from Figure \ref{fig:validation_metrics}.(a),(c) we 
show that, as it is desirable, the differences in settling time of pCTQL and CTQL with respect to Q-learning are not statistically significant.
Secondly, when using reward \ref{eq:reward_distance} we observed that the CTQL achieves the best (lowest) steady state error, whereas when using reward \ref{eq:reward_gym} (Figure \ref{fig:validation_metrics}.(d)), the smallest error is given by the pCTQL with $\omega = 0.01$.

\paragraph{Perturbed conditions.}
To test the robustness of the learned control strategies to changes in the environment, we generated $1000$ set-ups, by varying the initial conditions randomly (with a uniform distribution) in the state and control spaces, and varying the mass $m$ and length $l$ of the pendulum by $\pm 5\%$ of their nominal values, and using the Latin hypercube method \citep{loh1996latin}. The results, not portrayed here for the sake of brevity, show that, as is desirable, we obtain similar settling times for all the algorithms. Also, concerning the steady state error, when using reward \eqref{eq:reward_distance}, for all the algorithms, performance remain centered around that obtained under nominal conditions. Differently, when using reward \eqref{eq:reward_gym}, pCTQL displays a larger error when compared to that obtained under nominal condition (which was however lower than that of Q-learning), whereas Q-learning retains the same mean.

\section{Conclusions}
We presented a deterministic and a probabilistic Control-Tutored Q-learning strategy, that integrate a feedback control law synthesized on a partial model of the plant within a Q-learning framework to render the learning process faster and improving the performance of the learnt policies in achieving a control goal of interest.
We compared the control-tutored strategies with a classical Q-learning approach using the inverted pendulum stabilization benchmark from OpenAI Gym as a representative control problem.
We found that, when compared to Q-learning, CTQL requires fewer data samples and has a larger average reward, while pCTQL yields higher rewards with a comparable number of data samples; moreover, both CTQL and pCTQL yield lower regulation error when certain reward functions are used. Our numerical results show that both from a learning and a control viewpoint using a control-tutored learning approach might be beneficial. 

The next step is the derivation of proofs of convergence for the control-tutored algorithms presented in this paper.
Also, we wish to uncover and formally characterize the relationships  among the specific choice of the reward function, the performance of the algorithms and the approximate system dynamics needed to synthesize the control tutor.
We wish to emphasize that embedding a control tutor in the loop could be used to render more efficient learning strategies other than $Q$-learning. This will also be the subject of future investigation.
%

\newpage
\bibliography{refs.bib}

\end{document}